\documentclass[10pt,twocolumn,letterpaper]{article}

\usepackage[pagenumbers]{cvpr}      

\usepackage{float}
\usepackage[most]{tcolorbox}

\definecolor{cvprblue}{rgb}{0.21,0.49,0.74}
\usepackage[pagebackref,breaklinks,colorlinks,allcolors=cvprblue]{hyperref}

\usepackage{arydshln}
\usepackage{multirow}
\usepackage{float}

\usepackage{algorithm}
\usepackage{algorithmic}

\def\paperID{9441} 
\def\confName{CVPR}
\def\confYear{2026}

\title{Content-Aware Adaptive Token Pruning for Video-Language Models}

\author{
Paribesh Regmi\thanks{Work done during internship at Amazon.}, Qingshuang Chen, Chi Zhang, Heba Aly, Yelin Kim, Hongda Mao\\
Amazon.com Services LLC\\
{\tt\small paribesh@amazon.com}
}

\begin{document}
\maketitle
\begin{abstract}
Vision-language models excel at image and video understanding but suffer from high inference latency due to the need to process thousands of tokens per image, limiting their deployment on resource-constrained edge devices and in real-time surveillance applications. This challenge is further amplified in video processing, where multiple frames must be analyzed simultaneously. Existing token reduction techniques generally rely on a fixed, uniform pruning ratio applied across all inputs, which is suboptimal because the degree of redundancy can vary significantly between different videos, necessitating content-dependent pruning levels to preserve critical information. To address these limitations, we propose an adaptive token pruning strategy specifically designed for video processing. Crucially, the pruning ratio is determined adaptively based on the content of each video. This is achieved by analyzing the correlation structure of token embeddings to quantify redundancy, enabling dynamic token allocation that retains more tokens for dynamic scenes while aggressively pruning repetitive or static content. Importantly, our method is entirely post-hoc and requires no additional training or fine-tuning, while achieving strong empirical gains; notably, it improves accuracy by +7\% on a video captioning benchmark at 10\% token retention, while reducing computation TFLOPs by 95\%.
\end{abstract}    
\section{Introduction}
\label{sec:intro}
Vision-Language Models (VLMs) have demonstrated strong performance across image and video understanding tasks \cite{Zhang2024:Llava-Video, Zhu2025:Internvl3, Bai2025:qwen2_5vl, Li2024:llava-onevision, Dai2023:InstructBLIP}. The models encode visual inputs into token embeddings using a vision encoder and input them to a Large Language Model (LLM), together with textual instructions, to generate semantic descriptions of the visual content. Notably, encoding a single image can yield thousands of tokens, which imposes substantial computational overhead, as the processing cost of the LLM increases quadratically with the token sequence length. This challenge is further exacerbated in video processing, where multiple frames must be encoded and processed simultaneously. For example, in LLaVA-Video, each frame produces roughly 180 visual tokens; thus, a moderate 5-minute clip sampled at 5 fps already yields $\sim$11k tokens, making quadratic-time processing prohibitively expensive for low-resource edge devices. To address this, token reduction methods have been developed to reduce the number of tokens and feed only a fraction of them to the LLM decoder \cite{Alvar2025:DivPrune, Chen2024:FastV, Shang2024:PruMerge, Wang2024:DynamicVLM, Sun2025:llava-scissor, Liu2025:VideoXLPro, Huang2024:IVTP}.

Token reduction methods aim to retain only a subset of the most informative tokens from the visual input. One line of work trains a dedicated compressor or selector module to map the original tokens into a substantially smaller set \cite{Liu2025:VideoXLPro, Huang2024:IVTP}, but this approach requires additional training and cannot be seamlessly applied to pre-trained vision-language models. An alternative line of research develops training-free pruning strategies \cite{Alvar2025:DivPrune, Shang2024:PruMerge, Chen2024:FastV, Ye2025:ATP, Sun2025:llava-scissor}, which eliminate tokens without retraining. However, these methods predominantly impose a fixed, uniform pruning ratio across all inputs, failing to account for the varying degrees of redundancy across different videos. We argue that such one-size-fits-all strategies are suboptimal, and that adaptively allocating the token budget according to video content is essential for effective video understanding.

These approaches can be broadly categorized into attention-based pruning methods \cite{Shang2024:PruMerge, Chen2024:FastV, Ye2025:ATP} and embedding-based pruning methods \cite{Alvar2025:DivPrune, Sun2025:llava-scissor}, with the latter generally achieving superior performance. While these methods propose strategies for identifying representative tokens, most of them lack a principled mechanism for determining how many tokens are required to faithfully preserve visual information. This is critical, since different videos demand different length of tokens depending on the dynamism of their content. Using more tokens than necessary leads to inefficiency, whereas using fewer degrades performance. LLaVA-Scissor \cite{Sun2025:llava-scissor} merges semantically related components into a single token, effectively ending up with a different number of tokens for different inputs. However, its chain-merging procedure may conflate components where the endpoints exhibit weak or no semantic similarity, potentially leading to information loss. While prior pruning strategies rely on empirical heuristics such as attention weights or token norms, which recent studies show do not reliably reflect semantic importance \cite{Guo2024:AttentionIsNot}, a principled mechanism grounded in the intrinsic geometry and correlation of visual representations is needed to objectively quantify redundancy. Thus, we lack a principled method to reliably estimate the number of tokens necessary to represent the information content of a given visual input.

To address this challenge, we develop a novel adaptive token pruning strategy that quantitatively estimates redundancy to dynamically determine the retention ratio for each video. Specifically, we analyze the linear correlation structure among token embeddings by performing eigen-decomposition of their correlation matrix. The resulting eigenvalues capture the variance explained by different components, providing an objective measure of redundancy: if a few dominant eigenvalues explain most of the variance, the embeddings are highly redundant; conversely, a flatter distribution indicates greater diversity. To characterize this distribution, we fit the eigenvalue spectrum with an exponentially decaying function and extract its decay rate $\alpha$. A higher decay rate corresponds to stronger redundancy, implying that fewer tokens are sufficient to represent the visual input, while a slower decay indicates the need to retain more tokens. Accordingly, we define the retention ratio to be inversely proportional to the decay rate. For better empirical performance, this adaptive pruning is integrated within a two-stage frame- and token-level reduction pipeline that progressively discards redundant visual content.

In summary, our contributions are threefold:
\begin{itemize}
\item
We introduce a novel training-free adaptive token pruning method for video VLMs that dynamically determines the pruning ratio based on video content, eliminating rigid one-size-fits-all retention ratios.

\item
We formulate a principled spectral analysis of token correlations, modeling eigenvalue decay to quantitatively measure video redundancy and dynamically derive content-dependent retention budgets.

\item
We demonstrate consistent accuracy gains across three VLMs and five benchmark datasets under aggressive pruning (e.g., +7\% accuracy on VideoDC at 10\% retention while reducing TFLOPs by 95\%), showing that our adaptive pruning strategy preserves critical semantic fidelity while drastically cutting computation.
\end{itemize}

Unlike prior fixed-ratio pruning methods, our method determines the pruning ratio for each video based on its content.

\begin{figure*}  
  \centering
  \begin{subfigure}{0.39\linewidth}
    \includegraphics[width=\linewidth]{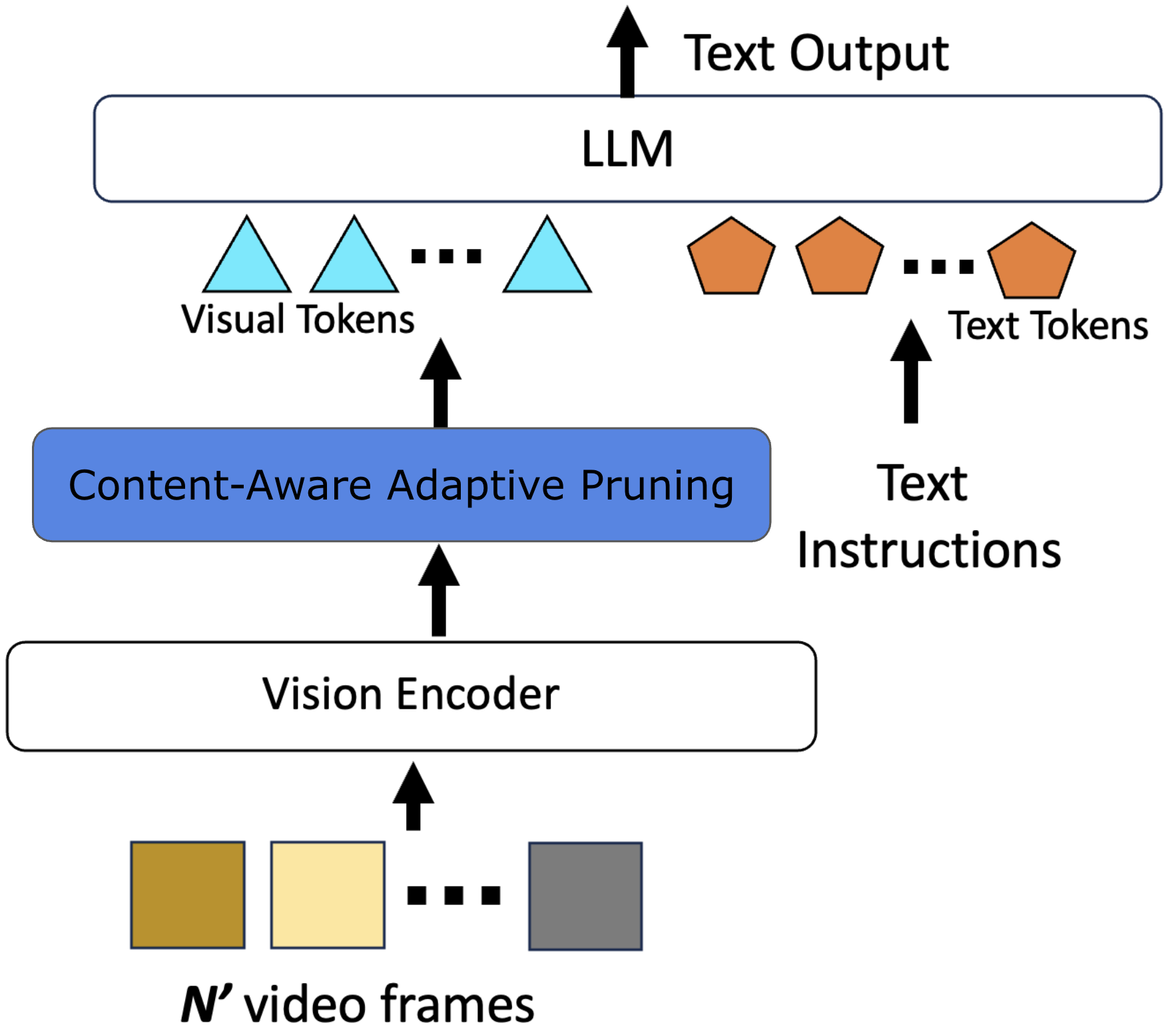}
    \caption{}
    \end{subfigure}
    \hspace{0.04\linewidth}
  \begin{subfigure}{0.55\linewidth}
    \includegraphics[width=\linewidth]{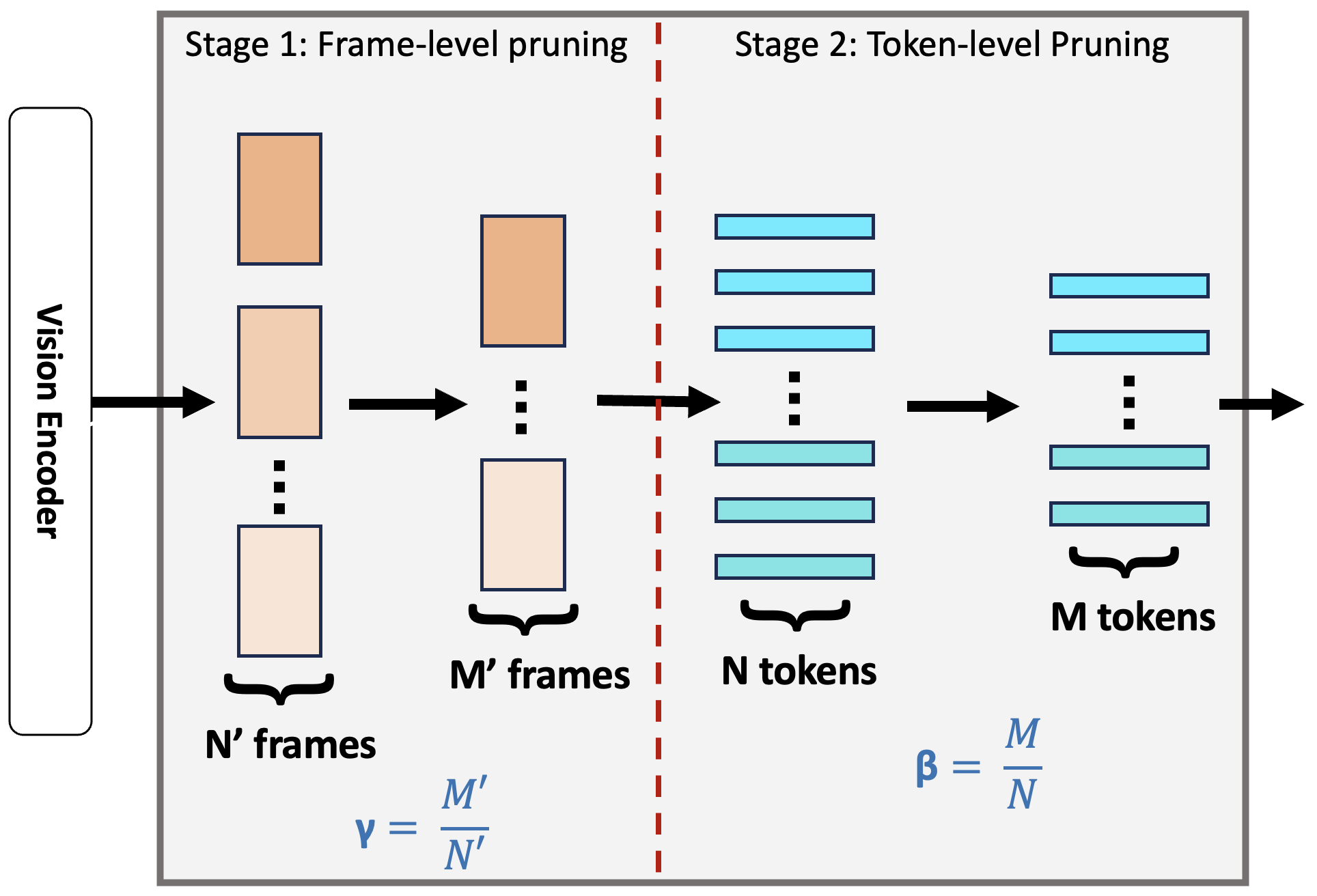}
    \caption{}
    \end{subfigure}
  \caption{(a) A general block diagram of a VLM with our pruning algorithm applied on top of the vision encoder. (b) Pruning pipeline: To improve performance, we employ diversity-based selection \cite{Alvar2025:DivPrune} to select $M'$ frames out of $N'$ frames in the first stage with retention ratio $\gamma$. In the second stage, our adaptive token pruning is applied where the retention ratio $\beta$ is determined based on the video content. The overall retained compute budget is therefore $\gamma$ (frame retention) $\times$ $\beta$ (token retention).}
  \label{fig:two_stage_prune}
\end{figure*}

\section{Related Works}
\label{sec:lit_review}
\subsection{Small-Scale VLMs}
Due to the high computational demands of large VLMs, recent research has focused on developing algorithms that enhance efficiency without sacrificing performance. One line of work proposes training strategies to boost the capabilities of VLMs with relatively small parameter sizes \cite{Zhou2024:Tinyllava, He2024:Bunny, Shao2025:Imp}. Another line of work propose knowledge distillation techniques, transferring knowledge from large-scale models to their smaller counterparts \cite{Shu2024:Llava-mod, cai2024:Llava-KD}. However, due to their small size, these models lose the powerful generation capability of the large-parameter models, and significantly fall behind in performance.

\subsection{Visual Token Reduction}
VLMs typically process thousands of visual tokens, and their computational complexity scales quadratically with input length. Thus, another line of research improves efficiency not by reducing model size, but by reducing the number of visual tokens \cite{Liu2025:VideoXLPro, Han2024:Filter, Shen2024:LongVU, Alvar2025:DivPrune, Sun2025:llava-scissor, Shang2024:PruMerge}. Video-XL-Pro \cite{Liu2025:VideoXLPro} trains a compressor module that compresses a large number of tokens into one. LongVU \cite{Shen2024:LongVU} and LLaMA-Vid \cite{Li2024:Llama-Vid} improve efficiency through query-based selection of salient tokens using attention layers. Other approaches, such as VideoLLaMB \cite{Wang2024:VideoLLamB} and VideoLLaMA2 \cite{Cheng2024:VideoLLamA2}, adopt token compression mechanisms, for example training an additional compressor module, via convolutional layers or memory bridges that aggregate multiple visual tokens into a single memory token. A key limitation of these methods is the need for training or fine-tuning additional compressor modules, which restricts their applicability to off-the-shelf VLMs.

There are training-free methods that can be directly used to reduce visual tokens in off-the-shelf VLMs \cite{Han2024:Filter, Alvar2025:DivPrune, Sun2025:llava-scissor, Shang2024:PruMerge, Chen2024:FastV, Wang2024:DynamicVLM, Yang2025:TopV, Ye2025:ATP}. Among these, PruMerge \cite{Shang2024:PruMerge}, ATP \cite{Ye2025:ATP}, and FastV \cite{Chen2024:FastV} reduce tokens by selecting only the ``important tokens" from all visual tokens, where importance is estimated based on the total attention a token receives in the self-attention layers of either the vision encoder or the LLM component of the VLM. However, it turns out that attention scores are not the sole indicators of token importance \cite{Guo2024:AttentionIsNot}. DivPrune \cite{Alvar2025:DivPrune} selects a fixed number of visual tokens by maximizing diversity, thereby retaining the least redundant subset. Diversity is measured via cosine similarity between token embeddings, and the method demonstrates improved performance over attention-based selection. However, it requires a pre-determined number of tokens (fixed retention ratio) for all inputs, adopting a one-size-fits-all strategy. This is suboptimal since inputs vary in redundancy, and the number of tokens needed to adequately represent information differs across cases. LLaVA-Scissor \cite{Sun2025:llava-scissor} extends this line of work with a graph-based token compression method. It constructs a graph where nodes represent tokens and edges connect similar tokens, then merges all tokens within each connected component into a single representative token. Unlike DivPrune, in LLaVA-Scissor the number of retained tokens adapts to each video, overcoming the fixed-token limitation. A limitation, however, arises from the transitivity of connectivity: in long chains of connected components, tokens at the ends may be grouped together despite being semantically dissimilar, leading to potential loss of distinct information. Also, the computation of connected components is in itself a computationally expensive operation.

To overcome these limitations, we propose an adaptive token pruning algorithm that dynamically determines the retention ratio for each video based on token correlation analysis.


\subsection{Correlation-based Pruning Methods}
Spectral and correlation-based criteria have long been used to identify redundancy for pruning and compression. Early work used principal-component analysis of activations to remove redundant parameters \cite{Levin1993:Fast}. More recently, heavy-tailed spectral analysis of weight matrices has been used to allocate layer-wise sparsity in large language models \cite{Lu2024:AlphaPruning}, and low-rank / SVD approaches have been applied to compress and prune weights while preserving model behavior. Moreover, the previously discussed methods, DivPrune \cite{Alvar2025:DivPrune} and LLaVA-Scissor \cite{Sun2025:llava-scissor}, also exploit correlations between token embeddings to either select or merge tokens. In this work, we leverage the correlation matrix as an indicator of redundancy, using it to estimate the number of tokens required to effectively represent the visual information content.
\section{Methodology}
\label{sec:method}
In this section, we present our content-adaptive token pruning methodology. We analyze the linear correlation structure among visual token embeddings through spectral decomposition to quantify redundancy, and dynamically determine the token retention ratio tailored to each individual video.

\begin{figure*}  
  \centering
  \begin{subfigure}{0.99\linewidth}
    \includegraphics[width=\linewidth]{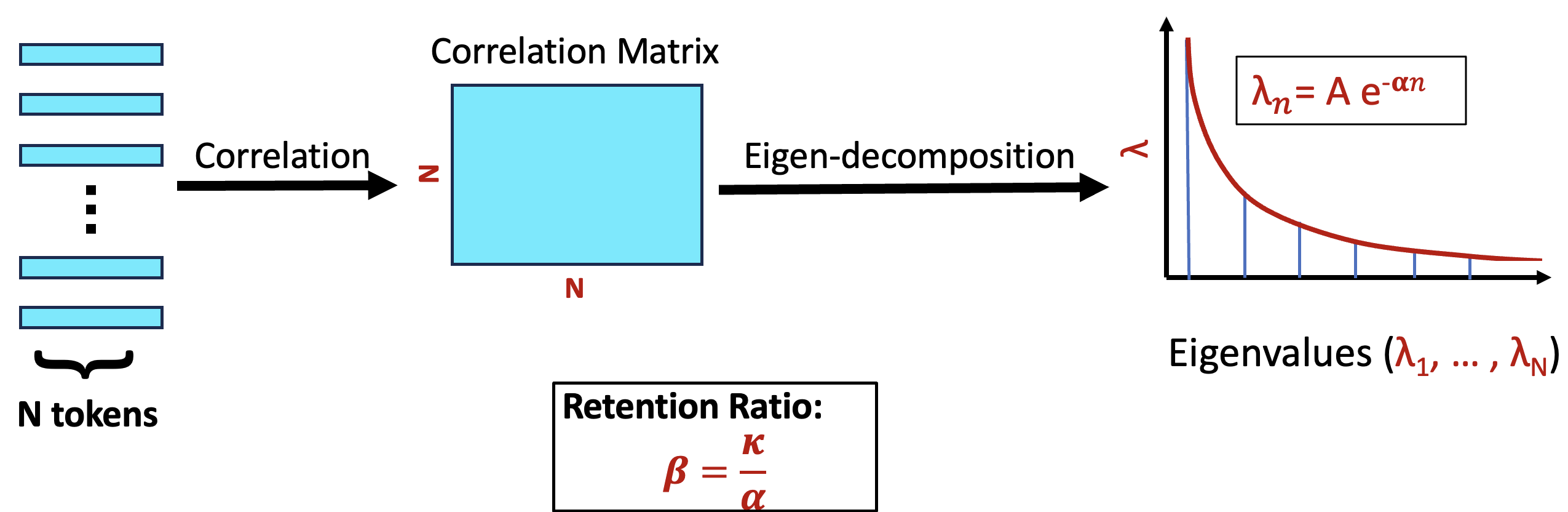}
    \end{subfigure}
  \caption{Determining the retention ratio $\beta$. After $N$ tokens are obtained from the first-stage pruning (Figure \ref{fig:two_stage_prune}(b)), pairwise correlations are computed to construct a correlation matrix. The eigenvalues of this matrix are then calculated, providing information about the degree of correlation (or redundancy) among the tokens. To quantify this, the eigenvalues are ordered in descending magnitude and an exponential curve is fitted to estimate the decay rate, which reflects the level of redundancy. Finally, the retention ratio is defined as inversely proportional to the decay rate.}
  \label{fig:determine_beta}
\end{figure*}

\begin{figure*}  
  \centering
  \begin{subfigure}{0.99\linewidth}
    \includegraphics[width=\linewidth]{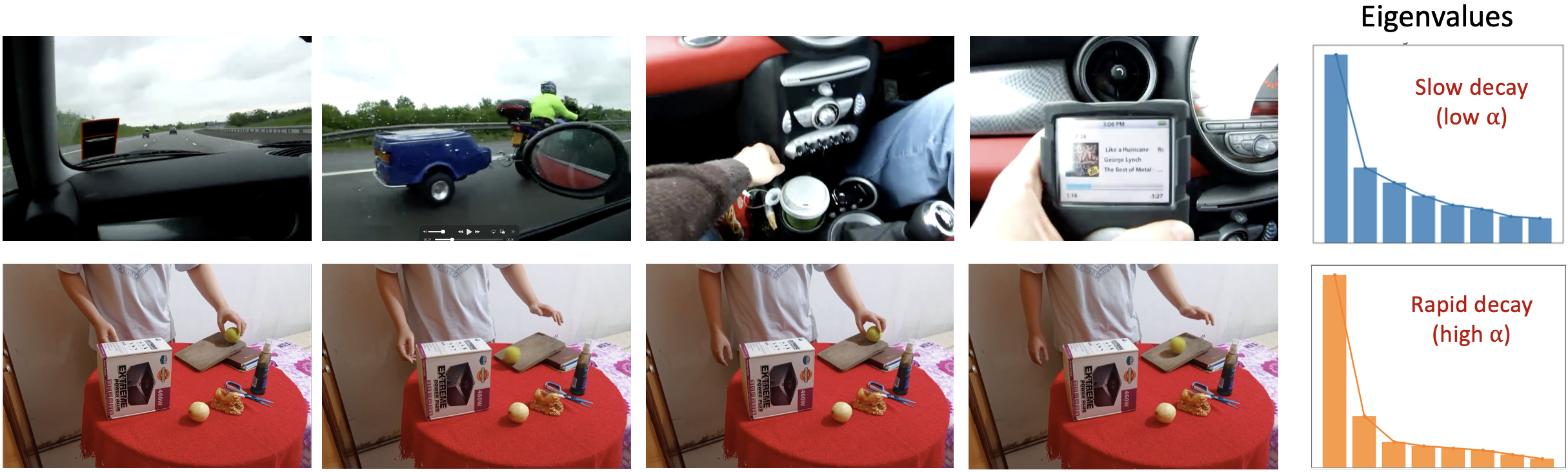}
    \end{subfigure}
  \caption{Demonstration of the eigenvalue spectrum for two representative videos. The top panel corresponds to a video captured from inside a moving car, featuring highly dynamic scenes, substantial camera motion, and multiple distinct activities across different scenes. The bottom panel shows a person rolling a ball down an incline, where the activity is repetitive throughout the video and the camera remains relatively static, focusing on a single scene. We perform eigen-decomposition of the token correlation matrix as described in Figure \ref{fig:determine_beta}, and the rightmost subfigure presents the corresponding eigenvalue spectra. For the dynamic video (top), the eigenvalues decay gradually, indicating that significant information is distributed across multiple modes. In contrast, for the static, repetitive video (bottom), the eigenvalues exhibit a steep decay, with most of the energy concentrated in the first eigenvalue}
  \label{fig:example}
\end{figure*}


\subsection{Adaptive Retention Ratio}
\label{sec:adaPrune}

Vision encoders such as CLIP and SigLIP, widely used in vision–language models, are trained with a contrastive objective that optimizes the cosine distance between visual embeddings and their corresponding textual descriptions. This training objective encourages the embeddings to preserve linear relationships in the embedding space: semantically similar patches exhibit high cosine similarity and vice versa. Thus, linear correlation between the visual embeddings informs about the redundancy in the visual information input to the model. We propose leveraging this information to estimate the number of visual tokens required to adequately represent the information in the input. A higher degree of correlation indicates greater redundancy, allowing the content to be represented with fewer tokens, whereas lower correlation suggests the need to retain more tokens to preserve information.

We propose to analyze redundancy in token embeddings through a linear decomposition of their correlation matrix. Specifically, we perform eigen-decomposition to obtain the eigenvalues, which capture the variance explained by each constituent component. The distribution of these values provides a quantitative measure of correlation among tokens: a small number of dominant principal values indicates high redundancy, whereas a more uniform distribution suggests greater diversity. We fit the sequence of eigenvalues with an exponentially decaying function, since exponential decay matches the empirically observed fast eigenvalue drop-off in redundant videos. This process is illustrated in Figure \ref{fig:determine_beta}. The rate of decay serves as an indicator of the degree of correlation in the tokens. Importantly, our framework is fully post-hoc and training-free, requiring no learnable parameters or fine-tuning. It explicitly decouples budget estimation from the token selection policy: our spectral correlation analysis is employed solely to determine the optimal token retention ratio $\beta$, which can then be paired with any token selection policy (we use diversity-based selection \cite{Alvar2025:DivPrune} in our primary implementation).

\begin{algorithm}[t]
\caption{Obtaining the adaptive retention ratio $\beta$}
\label{alg:beta}
\begin{algorithmic}[1]
\REQUIRE Token embeddings $\{\mathbf{x}_t\}_{t=1}^{T}$, scaling coefficient $\kappa$, number of required eigenvalues $K$.
\vspace{0.7em}
\STATE Obtain top-$K$ eigenvalues via low-rank SVD:
\vspace{-0.7em}
\[
\lambda_1, \ldots, \lambda_K = \texttt{SVD\_lowrank}(\{\mathbf{x}_t\}, K)
\]

\STATE Form the sequence of index–value data pairs:
\vspace{-0.7em}
\[
(n, \lambda_n) \text{ for } n=1,\ldots,K
\]

\STATE Estimate the decay rate by fitting the following curve to the data pairs:
\vspace{-0.7em}
\[
\log \lambda_n = \log A -\alpha n
\]

\STATE Compute the pruning ratio:
\vspace{-0.7em}
\[
\beta = \kappa / \alpha
\]

\RETURN $\beta$
\end{algorithmic}
\end{algorithm}

We begin by constructing the correlation matrix of the token embeddings and performing eigen-decomposition to obtain the eigenvalues $\lambda_1 \geq \lambda_2 \geq \cdots \geq \lambda_T$. These eigenvalues characterize the distribution of variance among the tokens: a spectrum dominated by a few large values indicates high redundancy, whereas a flatter spectrum reflects greater diversity. To quantify this redundancy, we model the eigenvalue spectrum with an exponential decay function:
\begin{align}
\lambda_n = A e^{-\alpha n}
\end{align}
where $n$ is the rank index and $\alpha$ is the decay rate. Taking the logarithm yields a log--linear relation,
\begin{align}
    \log \lambda_n = \log A - \alpha n,
\end{align}
which allows us to estimate $\alpha$ by fitting a straight line to $(n, \log \lambda_n)$. In practice, we use the top 32 eigenvalues, which capture most of the spectral energy and provide a stable estimate of the decay trend, while the smaller eigenvalues, which contribute little meaningful information, have minimal effect on the spectrum’s shape. A larger value of $\alpha$ indicates faster decay of the spectrum and thus higher redundancy among tokens, while a smaller value suggests greater diversity in the token representations as shown in Figure \ref{fig:example}. We employ a low-rank SVD (singular valued decomposition) approximation using iterative power methods to compute only the top 32 eigenvalues, providing a fast and memory-efficient alternative to full decomposition. This reduces the cubic time complexity of full decomposition to roughly quadratic in the matrix size, making it scalable for large token sets.

Based on this redundancy score, we define the pruning ratio as
\begin{align}
   \beta = \frac{\kappa}{\alpha} 
\end{align}
where $\kappa$ is a scaling factor that controls the overall pruning strength.  An algorithm for estimating $\beta$ is presented in Algorithm \ref{alg:beta}. Since $\alpha$ is derived from the eigenvalue spectrum of each video, the resulting pruning ratio $\beta$ is content-dependent and adapts automatically to the redundancy level of individual videos. Importantly, $\kappa$ is not tuned per video; it functions as a global compute-budget knob set by the system practitioner, analogous to specifying a target average compression rate or FLOPs constraint. While $\kappa$ dictates the overall operating point, per-video adaptivity is autonomously governed by the estimated spectral decay rate $\alpha$. In practice, $\kappa$ is calibrated once on a small representative set to match the desired average retention $\bar{\beta} = \kappa / \bar{\alpha}$ (where $\bar{\alpha}$ is the mean decay rate) and remains fixed during inference without requiring retuning for individual or unseen videos. This decay-rate-based pruning ratio serves as the basis for our adaptive token pruning framework.

\begin{table*}[t]
\centering
\small
\caption{Comparison of different pruning/retention methods across five video-language benchmarks. All baselines and our method share the same two-stage pipeline (frame-level selection followed by token-level pruning). 
VideoDC and Video-ChatGPT report both accuracy and score, while Next-QA, PerceptionTest, and Video-MME report accuracy only.}
\setlength{\tabcolsep}{4pt}
\begin{tabular}{clccccccc|cc}
\toprule
\multicolumn{1}{c}{\%Retention} & \multicolumn{1}{l}{\textbf{Method}} &
\multicolumn{2}{c}{\textbf{VideoDC}} & \multicolumn{2}{c}{\textbf{VideoChatGPT}} & 
\multicolumn{1}{c}{\textbf{Next-QA}} & \multicolumn{1}{c}{\textbf{PerceptionTest}} & 
\multicolumn{1}{c|}{\textbf{Video-MME}} & Time (ms) & TFLOPs \\
\cmidrule(lr){3-4} \cmidrule(lr){5-6} \cmidrule(lr){7-7} \cmidrule(lr){8-8} \cmidrule(lr){9-9}
 & & \textbf{Acc.} & \textbf{Score} & \textbf{Acc.} & \textbf{Score} & \textbf{Acc.} & \textbf{Acc.} & \textbf{Acc.} \\
 \midrule
 100\% & \textbf{Original} & 86.57 & 3.75 & 72.44 & 3.31 & 83.21 & 67.84 & 63.66 & 50.06 & 87.9 \\
\midrule
 & \textbf{PruMerge} & 65.33 & 3.07 & 54.74 & 2.65 & 77.58 & 59.61 & 57.33 & 42.30 \\
 & \textbf{FastV} & 51.10 & 2.66 & 50.22 & 2.54 & 75.11 & 56.97 & 50.59 & 41.30 \\
 & \textbf{AvgPool} & 78.15 & 3.46 & 65.88 & 3.05 & 80.26 & 63.55 & 59.44 & 39.90 \\
30\% & \textbf{DivPrune} & 80.96 & 3.55 & 69.43 & 3.18 & 82.05 & 65.05 & 60.48 & 41.44 &  14.9 \\
 & \textbf{LLaVA-Scissor} & 79.15 & 3.48 & 65.69 & 3.05 & 81.35 & 64.34 & 59.15 & 47.87 \\
 \noalign{\vskip 1pt}
\noalign{\vskip 1pt}
 & \textbf{Ours} & \textbf{83.56} & \textbf{3.61} & \textbf{70.31} & \textbf{3.21} & \textbf{82.15} & \textbf{65.93} & \textbf{60.52} & 42.19 \\
\midrule
 & \textbf{PruMerge} & 52.90 & 2.70 & 48.82 & 2.41 & 75.08 & 56.80 & 54.74 & 40.07 \\
 & \textbf{FastV} & 40.48 & 2.35 & 47.00 & 2.42 & 73.68 & 56.39 & 49.11 & 40.97 \\
 & \textbf{AvgPool} & 66.73 & 3.10 & 55.15 & 2.68 & 77.23 & 61.23 & 56.00 & 39.30 \\
15\% & \textbf{DivPrune} & 76.15 & 3.40 & 65.21 & 3.02 & 80.88 & 63.70 & 58.85 & 39.30 &  6.3  \\
 & \textbf{LLaVA-Scissor} & 65.13 & 3.09 & 56.52 & 2.74 & 79.60 & 62.71 & 56.33 & 42.38 \\
 \noalign{\vskip 1pt}
\noalign{\vskip 1pt}
 & \textbf{Ours} & \textbf{82.36} & \textbf{3.58} & \textbf{67.41} & \textbf{3.09} & \textbf{81.23} & \textbf{64.46} & \textbf{60.07} & 39.37 \\
\midrule
 & \textbf{PruMerge} & 36.47 & 2.15 & 39.31 & 2.04 & 70.96 & 53.55 & 51.07 & 39.2 \\
 & \textbf{FastV} & 34.66 & 2.13 & 43.73 & 2.31 & 73.01 & 56.17 & 49.04 & 39.86 \\
 & \textbf{AvgPool} & 52.30 & 2.71 & 47.54 & 2.39 & 73.77 & 56.85 & 53.37 & 39.0 \\
10\% & \textbf{DivPrune} & 68.47 & 3.19 & 59.54 & 2.84 & 79.49 & 61.55 & 57.85 & 38.7 & 3.9 \\
 & \textbf{LLaVA-Scissor} & 60.32 & 2.91 & 54.20 & 2.63 & 78.67 & 61.52 & 55.74 & 42.30 \\
 \noalign{\vskip 1pt}
\noalign{\vskip 1pt}
 & \textbf{Ours} & \textbf{76.75} & \textbf{3.41} & \textbf{63.12} & \textbf{2.95} & \textbf{79.95} & \textbf{62.62} & \textbf{57.88} & 38.87 \\
\bottomrule
\end{tabular}
\label{tab:performance1}
\end{table*}

\section{Experiments}

We conduct a series of experiments to assess the effectiveness of our proposed method.  First, we evaluate the performance of our method at different retention ratios by comparing it against the state-of-the-art training-free pruning baselines for the LLaVA-Video \cite{Zhang2024:Llava-Video} model. Next, we extend this comparison to other popular VLMs: InternVL3 \cite{Zhu2025:Internvl3} and Qwen2.5VL \cite{Bai2025:qwen2_5vl}. Then, we assess how the performance generalizes to different parameter sizes of the VLMs. Finally, we conduct ablation studies to isolate the impact of our content-adaptive retention ratio and evaluate different functional forms for modeling spectral decay. All experiments are conducted on NVIDIA L40S GPU with 48GB memory.

\subsection{Performance Comparison on LLaVA-Video}

We compare the performance of baselines and our method on the LLaVA-Video \cite{Zhang2024:Llava-Video} model with 7B parameters. 

\textbf{Video Datasets:} The models are evaluated on video captioning dataset VideoDetailCaption (VideoDC) \cite{LMMsLab:VideoDC}, open-ended QA dataset VideoChatGPT \cite{Maaz2023:VideoChatGPT}, and three multiple choice datasets NextQA \cite{Xiao2021:NextQA}, PerceptionTest \cite{Patraucean2023:PerceptionTest}, and Video-MME \cite{Fu2025:VideoMME}. 

\textbf{Baselines:} We consider five training-free token pruning baselines: PruMerge \cite{Shang2024:PruMerge}, FastV \cite{Chen2024:FastV}, AvgPool from Dynamic-VLM \cite{Wang2024:DynamicVLM}, DivPrune \cite{Alvar2025:DivPrune}, and LLaVA-Scissor \cite{Sun2025:llava-scissor}. To ensure a fair and consistent evaluation, all baselines are applied within the exact same two-stage framework as our method in Figure \ref{fig:two_stage_prune}: the second stage carries out token pruning using each respective baseline. We compare the models for 30\%, 15\% and 10\% retention ratios. For our method, since the retention ratio is different for different videos, we set the hyperparameter $\kappa$ such that on average the ratio is the same as baselines. Because our pruning ratio is content-adaptive, the actual token budget allocated varies dynamically across individual videos. For instance, at the 30\% average retention setting on LLaVA-Video, the retained ratio exhibits an empirical distribution of $30 \pm 6\%$, dynamically allocating larger token budgets to videos with complex, fast-changing scenes while aggressively compressing static or repetitive clips.

\textbf{Metrics:} For video captioning and open-ended QA datasets, we report the percentage accuracy and score (out of 5) of the predictions. These are obtained by prompting an LLM (Claude sonnet 3.7) to rate the prediction as compared to the ground truth answer. For multiple choice datasets, we report the percentage accuracy of the predictions.

\begin{table*}[t]
\centering
\caption{Comparison of different pruning/retention methods across three video-language benchmarks, with separate blocks for InternVL and QwenVL. VideoDC and VideoChatGPT report both accuracy and score, while Next-QA reports accuracy only.}
\small
\setlength{\tabcolsep}{4pt}
\begin{tabular}{clccccc|ccccc}
\toprule
\multicolumn{2}{c}{} & \multicolumn{5}{c}{\textbf{InternVL3}} & \multicolumn{5}{c}{\textbf{Qwen2.5VL}} \\
\cmidrule(lr){3-7} \cmidrule(lr){8-12}
\multicolumn{1}{c}{\%Retention} & \multicolumn{1}{l}{\textbf{Method}} &
\multicolumn{2}{c}{\textbf{VideoDC}} & \multicolumn{2}{c}{\textbf{VideoChatGPT}} & \multicolumn{1}{c}{\textbf{Next-QA}} &
\multicolumn{2}{c}{\textbf{VideoDC}} & \multicolumn{2}{c}{\textbf{VideoChatGPT}} & \multicolumn{1}{c}{\textbf{Next-QA}} \\
\cmidrule(lr){3-4} \cmidrule(lr){5-6} \cmidrule(lr){7-7} \cmidrule(lr){8-9} \cmidrule(lr){10-11} \cmidrule(lr){12-12}
 & & \textbf{Acc.} & \textbf{Score} & \textbf{Acc.} & \textbf{Score} & Accuracy & \textbf{Acc.} & \textbf{Score} & \textbf{Acc.} & \textbf{Score} & Accuracy \\
\midrule
& \textbf{PruMerge} & 71.34 & 2.75 & 55.74 & 2.75 & 79.50  & 79.35 & 3.44 & 64.31 & 3.11 & 81.60 \\
30\% & \textbf{DivPrune} & 75.35 & 3.26 & 57.3 & 2.78 & 81.76 & 77.95 & 3.43 & 65.84 & 3.15 & 81.84 \\
 & \textbf{LLaVA-Scissor} & 74.14 & 3.22 & 56.97 & 2.78 & 81.79 & 78.15 & 3.38 & 62.89 & 3.04 & 81.34\\
 \noalign{\vskip 1pt}
\noalign{\vskip 1pt}
 & \textbf{Ours} & \textbf{80.76} & \textbf{3.41} & \textbf{60.52} & \textbf{2.86} & \textbf{81.98} & \textbf{79.36} & \textbf{3.44} & \textbf{65.92} & \textbf{3.16} & \textbf{81.89} \\
\midrule
 & \textbf{PruMerge} & 63.73 & 2.97 & 51.06 & 2.58 & 76.12  & \textbf{77.75} & 3.40 & 64.38 & 3.09 & 80.19 \\
15\% & \textbf{DivPrune} & 72.34 & 3.14 & 55.55 & 2.71 & 80.77 & 75.75 & 3.37 & \textbf{65.95} & \textbf{3.14} & 80.85 \\
 & \textbf{LLaVA-Scissor} & 69.54 & 3.09 & 55.59 & 2.69 & 80.85 & 65.53 & 3.01 & 58.36 & 2.83 & 79.40\\
 \noalign{\vskip 1pt}
\noalign{\vskip 1pt}
 & \textbf{Ours} & \textbf{76.75} & \textbf{3.37} & \textbf{57.73} & \textbf{2.78} & \textbf{81.85} & 76.55 & 3.38 & 65.56 & 3.13 & \textbf{80.88} \\
\midrule
 & \textbf{PruMerge} & 56.91 & 3.21 & 46.24 & 2.43 & 74.45  & \textbf{76.35} & \textbf{3.37} & 63.83 & 3.07 & 79.15 \\
10\% & \textbf{DivPrune} & 64.13 & 2.98 & 53.35 & 2.64 & 79.57 & 73.34 & 3.26 & \textbf{64.35} & 3.06 & 80.07 \\
 & \textbf{LLaVA-Scissor} & 69.53 & 3.08 & 55.10 & 2.67 & 80.44 & 54.71 & 2.71 & 53.68 & 2.63 & 76.98 \\
 \noalign{\vskip 1pt}
\noalign{\vskip 1pt}
 & \textbf{Ours} & \textbf{75.15} & \textbf{3.26} & \textbf{55.72} & \textbf{2.69} & \textbf{81.48} & 75.15 & 3.33 & \textbf{64.36} & \textbf{3.08} & \textbf{80.18} \\ 
\bottomrule
\end{tabular}
\label{tab:performance2}
\end{table*}

The results are reported in Table \ref{tab:performance1}. For all the baselines and our method, we observe a decrease in performance as the retention ratio decreases. Across all retention ratios, our method consistently outperforms the baseline. Furthermore, we observe that in comparison with the multiple-choice QA tasks, the performance gain with our method is more significant in captioning (VideoDC) and the open-ended QA (VideoChatGPT) tasks. Specially for low retention ratios (15\% and 10\%), we achieve over 6\% improvement on VideoDC, and 2-4\% improvement on VideoChatGPT over the baselines.

The last two columns of the table report the time per token generation on the VideoDC dataset and the number of floating point operations in LLM for each retention ratio, measured in trillions (TFLOPs). The reported generation time is an end-to-end measurement per token that accounts for all operational stages, including low-rank SVD computation, token correlation analysis, pruning, and the LLM forward pass. On average, 250 tokens are generated by each method, and the per-token time is calculated accordingly. It is observed that LLaVA-Scissor requires substantially more time than other methods, as it computes all connected components of the similarity graph—an inherently expensive operation. In contrast, our method is considerably more efficient. Although the token matrices have ranks on the order of thousands, only a small number of eigenvalues (32 in our case) are required to estimate the decay rate $\alpha$, which allows for low-rank SVD estimation, making the computation significantly cheaper than performing a full eigen-decomposition (computing the top-32 eigenvalues via low-rank SVD takes $< 1.5$ ms per video, introducing negligible overhead relative to the total inference latency). From the last column, we observe that there is $\sim83\%$, $\sim93\%$, and $\sim95.5\%$ reduction in TFLOPs at 30\%, 15\%, and 10\% retention ratios. Notably, at 30\% retention in VideoDC and VideoChatGPT datasets, our method only suffers 2-3\% drop in performance, while lowering the computational overhead by 83\%.

\subsection{Extending Comparison to InternVL3 and Qwen2.5VL}

In Table \ref{tab:performance2}, we extend the comparison of our method and baselines to other state-of-the-art vision-language models: InternVL3 \cite{Zhu2025:Internvl3} and Qwen2.5VL \cite{Bai2025:qwen2_5vl} of 8B and 7B parameter sizes respectively. For comparison, we use the best baselines from Table \ref{tab:performance1} (Prumerge, DivPrune, and LLaVA-Scissor), and datasets from all 3 categories i.e. captioning (VideoDC), open-ended QA (VideoChatGPT) and multiple-choice QA (NextQA). The results show that our method achieves the overall best performance, which is consistent with Table \ref{tab:performance1}. This demonstrates the versatility of our adaptive pruning approach across different VLMs.

\subsection{Performance Across Different Parameter Sizes of VLMs}
We analyze if the performance of our method generalizes across different parameter sizes of the VLM. To this end, we compare the performance of our method and DivPrune, which is the overall best baseline from Tables \ref{tab:performance1} \& \ref{tab:performance2} on InternVL3 with 1B and 2B parameter sizes. Table \ref{tab:parameter_sizes} reports results on the video captioning (VideoDC) and multiple choice QA (NextQA) datasets, and our method has consistent and significant improvement over DivPrune across all retention ratios. Similar to table \ref{tab:performance1}, the performance gain is considerably higher at lower retention ratios, and for the video captioning task as compared to the multiple choice QA task. This further demonstrates the versatility of our proposed method across different parameter sizes of the VLM.

\subsection{Ablation Study}

\begin{table}[t]
\centering
\caption{Comparison of DivPrune and Ours on VideoDC and Next-QA across InternVL model sizes (1B and 2B). Results are reported in accuracy (\%) and score for VideoDC and accuracy for NextQA.}
\label{tab:parameter_sizes}
\begin{tabular}{clcc}
\toprule
\multicolumn{2}{c}{} & \textbf{InternVL-1B} & \textbf{InternVL-2B} \\
\bottomrule
\multicolumn{4}{l}{\textbf{VideoDC (Acc./Score)}} \\
\bottomrule
30\% & \textbf{DivPrune} & 62.53/2.89 & 71.54/3.16 \\
     & \textbf{Ours}     & \textbf{70.74/3.13} & \textbf{73.54/3.17} \\
\noalign{\vskip 2pt}
\noalign{\vskip 2pt}
15\% & \textbf{DivPrune} & 64.53/2.95 & 71.94/3.13 \\
     & \textbf{Ours}     & \textbf{68.13/3.05} & \textbf{75.15/3.18} \\
\noalign{\vskip 2pt}
\noalign{\vskip 2pt}
10\% & \textbf{DivPrune} & 61.52/2.85 & 65.73/3.02 \\
     & \textbf{Ours}     & \textbf{67.53/3.07} & \textbf{74.14/3.22} \\
\bottomrule
\multicolumn{4}{l}{\textbf{Next-QA (Acc.)}} \\
\bottomrule
30\% & \textbf{DivPrune} & 68.03 & 73.08 \\
     & \textbf{Ours}     & \textbf{69.12} & \textbf{74.25} \\
\noalign{\vskip 2pt}
\noalign{\vskip 2pt}
15\% & \textbf{DivPrune} & 67.21 & 71.99 \\
     & \textbf{Ours}     & \textbf{69.29} & \textbf{73.51} \\
\noalign{\vskip 2pt}
\noalign{\vskip 2pt}
10\% & \textbf{DivPrune} & 66.76 & 70.84 \\
     & \textbf{Ours}     & \textbf{68.65} & \textbf{73.36} \\
\bottomrule
\end{tabular}
\end{table}

\begin{table}[t]
\centering
\caption{Ablation study on the adaptive retention ratio. We compare our adaptive retention ratio against a fixed (uniform) retention ratio under the same reduction pipeline. Across both VideoDC and VideoChatGPT, our content-adaptive method consistently outperforms fixed pruning, with the largest gains observed at aggressive retention budgets.}
\label{tab:adaptive_ablation}
\begin{tabular}{ccc}
\toprule
 & \textbf{Fixed Ratio} & \textbf{Adaptive (Ours)} \\
\midrule
\multicolumn{3}{l}{\textbf{VideoDC (Acc./Score)}} \\
\midrule
30\% & 82.16/3.60 & \textbf{83.56/3.61}\\
15\% & 77.35/3.42 & \textbf{82.36/3.58} \\
10\% & 69.27/3.19 & \textbf{76.75/3.41} \\
\midrule
\multicolumn{3}{l}{\textbf{VideoChatGPT (Acc./Score)}} \\
\midrule
30\% & 69.96/3.21 & \textbf{70.31/3.21} \\
15\% & 66.55/3.08 & \textbf{67.41/3.09} \\
10\% & 60.94/2.91 & \textbf{63.12/2.95} \\
\bottomrule
\end{tabular}
\end{table}

\begin{table}[t]
\centering
\caption{Ablation study on the choice of functional form for fitting the eigenvalue spectrum. We compare linear, quadratic, and exponential decay rates to obtain the adaptive retention ratio $\beta$. Results are reported on VideoDC dataset across different retention ratios. The exponential formulation demonstrates superior performance.}
\label{tab:curve_ablation}
\begin{tabular}{ccccc}
\toprule
 Ret'n &  & \textbf{Linear} & \textbf{Quadratic} & \textbf{Exponential} \\
\midrule
30\% 
 & Acc.  & 82.35 & 82.36 & \textbf{83.56} \\ 
 & Score & \textbf{3.61}  & 3.59  & \textbf{3.61}  \\
\midrule
15\% 
 & Acc.  & 79.75 & 79.35 & \textbf{82.36} \\ 
 & Score & 3.49  & 3.50  & \textbf{3.58} \\
\midrule
10\% 
 & Acc.  & 74.34 & 75.35 & \textbf{76.75} \\ 
 & Score & 3.35  & 3.38  & \textbf{3.41} \\
\bottomrule
\end{tabular}
\end{table}

\subsubsection{Ablation on Adaptive Retention Ratio}
To validate the effectiveness of our proposed content-adaptive retention ratio, we ablate it against a fixed-ratio baseline that applies a uniform pruning budget across all videos under the same reduction pipeline. As shown in Table \ref{tab:adaptive_ablation}, our adaptive pruning strategy consistently outperforms the fixed-ratio counterpart across all retention levels on both VideoDC and VideoChatGPT. Notably, at aggressive pruning budgets (15\% and 10\% retention), adaptive pruning achieves significant gains (+5.01\% and +7.48\% accuracy on VideoDC, respectively). This demonstrates that dynamically allocating tokens according to video redundancy is crucial for preserving vital visual information in dynamic scenes while eliminating redundancy in static videos, rendering static one-size-fits-all pruning suboptimal.

\subsubsection{Ablation on Decay Rates}

In Section~\ref{sec:adaPrune}, we introduced the use of an exponential fit to the eigenvalue spectrum in order to obtain an exponential decay rate $\alpha$. In this section, we extend the analysis by exploring alternative functional forms for modeling the decay, specifically linear and quadratic curves.
\begin{align}
    \lambda_n = A - \alpha n \quad \text{(Linear)}\\
    \lambda_n = A - \alpha n^2 \quad \text{(Quadratic)}
\end{align}
For each formulation, the decay rate $\alpha$ is used to derive the adaptive retention ratio $\beta$. Table~\ref{tab:curve_ablation} reports the comparative results. We observe that  the exponential formulation achieves superior results overall, indicating that it provides a more faithful characterization of redundancy in the token embeddings.

\section{Conclusion}
We proposed a content-aware adaptive visual token pruning strategy for video inputs in vision-language models. Our adaptive strategy involves analyzing the linear correlation between the token embeddings to estimate the retention ratio for a video. Experiments show that the proposed method outperforms the training-free baselines across different VLMs and parameter sizes, demonstrating its effectiveness and versatility. An interesting future exploration is to study the effect of fine-tuning VLM with the proposed strategy on performance.

{
    \small
    \bibliographystyle{ieeenat_fullname}
    \bibliography{main}
}

\clearpage
\setcounter{page}{1}
\maketitlesupplementary

\section{Hyperparameters for Our Method}

For generation with all methods, we select the token with highest probability score from the LLM output. Also, we use beam size of 1, and the number of maximum new tokens is capped to 1024. A total of 64 frames were sampled from the videos for LLaVA-Video and InternVL following the original implementation. For QwenVL, we follow the original implementation to sample 2 frames per token, but limit the maximum frames to 64. We found out that $\gamma=0.5$ was consistently better performing in all settings for LLaVA-Video and InternVL, whereas for QwenVL, we set $\gamma=1$. The setting of hyperparameter $\kappa$ for our method is detailed in Table \ref{tab:kappa}.

\begin{table*}[!h]
    \centering
    \begin{tabular}{cccccc}
    \toprule
    Retention Ratio & \textbf{VideoDC} & \textbf{VideoChatGPT} & \textbf{NextQA} & \textbf{PerceptionTest} & \textbf{VideoMME} \\
    \midrule
    \multicolumn{3}{l}{LLaVA-Video} & & \\
    \midrule
    30\% & 0.064 & 0.064 & 0.0185 & 0.0465 & 0.055\\
    15\% & 0.032 & 0.032 & 0.074 & 0.023 & 0.027\\
    10\% & 0.016 & 0.016 & 0.037 & 0.0115 & 0.015\\
    \midrule
    \multicolumn{3}{l}{InternVL3} & & \\
    \midrule
    30\% & 0.0468 & 0.0468 & 0.0170 & - & - \\
    15\% & 0.0234 & 0.0234 & 0.0255 & - & - \\
    10\% & 0.0145 & 0.0145 & 0.0500 & - & - \\
    \midrule
    \multicolumn{3}{l}{Qwen2.5VL} & & \\
    \midrule
    30\% & 0.009 & 0.009 & 0.009 & - & - \\
    15\% & 0.0135 & 0.0135 & 0.0135 & - & - \\
    10\% & 0.0270 & 0.0270 & 0.0270 & - & - \\
    \bottomrule
    \end{tabular}
    \caption{The values of hyperparameter $\kappa$ used for our method on different VLMs for various retention ratios.}
    \label{tab:kappa}
\end{table*}

\section{Prompt Used for Evaluation}
We queried the LLM Claude Sonnet 3.7 with the following prompt to evaluate the generated response from VLMs. This prompt is copied from the LLaVA-NeXT repo (https://github.com/LLaVA-VL/LLaVA-NeXT):

\begin{tcolorbox}[
    colback=gray!10,
    colframe=black,
    boxrule=0.5pt,
    sharp corners,
    enhanced,
    width=\linewidth
]
You are an intelligent chatbot designed for evaluating the correctness of generative outputs for question-answer pairs.
        Your task is to compare the predicted answer with the correct answer and determine if they match meaningfully. Here's how you can accomplish the task

        \#\#INSTRUCTIONS:
        
        - Focus on the meaningful match between the predicted answer and the correct answer.
        
        - Consider synonyms or paraphrases as valid matches.
        
        - Evaluate the correctness of the prediction compared to the answer.

        Please evaluate the following video-based question-answer pair:
        
        Question: \{question\}
        
        Correct Answer: \{answer\}
        
        Predicted Answer: \{pred\}

        Provide your evaluation only as a yes/no and score where the score is an integer value between 0 and 5, with 5 indicating the highest meaningful match.
        Please generate the response in the form of a Python dictionary string with keys 'pred' and 'score', where value of 'pred' is a string of 'yes' or 'no' and value of 'score' is an INTEGER, not STRING.
        DO NOT PROVIDE ANY OTHER OUTPUT TEXT OR EXPLANATION. Only provide the Python dictionary string.
        For example, your response should look like this: \{\{"pred": "yes", "score": 4\}\}.
\end{tcolorbox}

\end{document}